\documentclass[accepted,specialissue]{melba}

\usepackage{amsmath,amsfonts,cleveref,subfigure,float,bbm,multirow,makecell,algpseudocode,algorithm}

\melbaid{2026:027}  
\doi{10.59275/j.melba.2026-c874}
\melbaauthors{Laufer, Mairhöfer, Sieren, Gerdes, Leal dos Reis, Bischof, Käster, Barth, Barkhausen, and Martinetz}  
\email{m.laufer@uni-luebeck.de}
\volume{2026}
\firstpageno{553}  
\melbayear{2026}  
\datesubmitted{2025-12}  
\datepublished{2026-07}  

\melbaspecialissue{Medical Imaging with Deep Learning (MIDL) 2025}
\melbaspecialissueeditors{Lisa Koch, Ronald M. Summers, Chen Chen, Yan Zhuang}

\ShortHeadings{Patient Pose Assessment Using a CT-Based Framework for Synthetic Data Generation}{Laufer, Mairhöfer, Sieren, Gerdes, Leal dos Reis, Bischof, Käster, Barth, Barkhausen, and Martinetz}

\title{Patient Pose Assessment Using a CT-Based Framework for Synthetic Data Generation}

\author{
	\firstname Manuel \surname Laufer\aff{1},
	\firstname Dominik \surname Mairhöfer\aff{1},
	\firstname Malte \surname Sieren\aff{2},
	\firstname Hauke \surname Gerdes\aff{2},
	\firstname Fabio \surname Leal dos Reis\aff{2},
	\firstname Arpad \surname Bischof\aff{2,3},
	\firstname Thomas \surname Käster\aff{4},
	\firstname Erhardt \surname Barth\aff{1},
	\firstname Jörg \surname Barkhausen\aff{2},
	\firstname Thomas \surname Martinetz\aff{1}}

\affiliations{
	\num 1 \addr Institute for Neuro- and Bioinformatics, University of Lübeck, Germany \\
	\num 2 \addr University Medical Center Schleswig-Holstein, Lübeck, Germany \\
	\num 3 \addr IMAGE Information Systems Europe GmbH, Rostock, Germany \\
	\num 4 \addr Pattern Recognition Company GmbH, Lübeck, Germany
}

\abstract{
An adequate diagnostic quality of radiographs is essential for reliable diagnoses and treatment planning. 
The patient's pose during radiography is one of the most important factors determining the diagnostic quality. 
Since patient positioning is difficult and not standardized, an automated AI-based approach using depth images to automatically assess the patient's pose before the radiograph has been taken would be helpful.
Due to regulatory hurdles, however, it is difficult in practice to acquire the required depth images and corresponding radiographs.
In this paper, we present a framework that can generate such training data synthetically from Computed Tomography scans. 
We further show that by pretraining on our generated synthetic dataset consisting of 3077 image pairs of upper ankle joints, the pose assessment of real upper ankle joints can be improved by up to 11 percentage points. 
}


\begin{document}

\twocolumn[\maketitle]

\section{Introduction}
\enluminure{T}{he} diagnostic quality of radiographs is essential for making reliable diagnoses and planning treatments.
Radiographs of inadequate diagnostic quality often lead to retakes and thus to increased radiation exposure for the patient and increased costs for the hospital.
In the worst case, inadequate diagnostic quality can lead to incorrect treatment and misdiagnosis.
The most important factor affecting the diagnostic quality of a radiograph is the pose of the patient at the time the radiograph is taken \citep{littleUnifiedDatabaseRejected2017}.
Furthermore, patient positioning is error-prone, as it is not standardized and depends heavily on the patient and the experience of the radiographer, who is also often under time pressure.

To assist the radiographer in positioning the patient and to protect the patient from increased radiation dose, an automatic pose assessment would help.
By attaching two Time-of-Flight (ToF) cameras to the X-ray device, we were able to show recently that depth images of anatomical preparations of upper ankle joints contain information that can lead to high accuracy pose assessment \citep{lauferPatientPoseAssessment2024}.
In order to determine a correspondence between the depth image of the pose and the diagnostic quality of the radiograph, the radiograph and the depth image must be taken simultaneously and labeled with their diagnostic quality. 
The depth image and the label can then be used to train neural networks to predict the diagnostic quality of the radiograph before the radiograph is even taken. 
However, radiographing subjects without an indication is problematic.
In particular, intentionally radiographing subjects in non-diagnostic poses, which are necessary for the training, is ethically difficult to justify.
Furthermore, using cameras in live clinical practice is not readily possible for data protection and regulatory reasons.
Finally, working with anatomical preparations as a solution is not scalable.

To address these challenges, we present a framework that synthetically generates the required image pairs of depth images and radiographs from Computed Tomography (CT) scans.
CT scans that have already been acquired can thus be used retrospectively to create large realistic synthetic datasets without additional radiation exposure, which makes the approach substantially more scalable than prospective data acquisition.
It is furthermore possible to generate multiple poses of varying diagnostic quality, including non-diagnostic ones, from a single CT scan by selectively adjusting the CT, further increasing the scalability and flexibility of the approach.
We show that by pretraining on our generated synthetic dataset of upper ankle joints, the pose assessment of real upper ankle joints from anatomical preparations can be improved by up to 11 percentage points (pp).
To further evaluate the transferability of this synthetic pretraining to a more realistic clinical setting, we have acquired an additional dataset of 18 living subjects reflecting the clinical practice of an X-ray examination, and we show that pretraining on the synthetic dataset can also improve performance on this realistic dataset.
The contribution of this work lies in the task-specific, clinically realistic design of the framework and in demonstrating that the generated synthetic data is useful for transfer to real datasets.
The synthetic dataset and the novel acquired realistic dataset is published under \url{https://github.com/INB-KI-SIGS/patient-pose-assessment.git}.
\section{Related Work}
For the generation of synthetic radiographs from CT scans, known as digitally reconstructed radiographs (DRR), two main approaches are used:
The ray tracing approach, using forward projection, casts a ray through the CT volume for each pixel on the detector and accumulates the intensity of the values along the path.
Although ray tracing is computationally efficient, it is not capable of modeling scattering or beam hardening \citep{russakoffFastGenerationDigitally2005}.
In the approach presented in \citet{unberathDeepDRRCatalystMachine2018}, a radiograph generated by forward projection is combined with deep learning-based scatter and noise estimation.
In contrast, the Monte Carlo (MC) approach simulates the transport of photons across the CT scan to model the photon-matter interaction and therefore requires material properties for each CT voxel \citep{badalAcceleratingMonteCarlo2009}.
Such simulations result in realistic DRRs; however, they are computationally more expensive than forward projection. 

Although there is little research on generating synthetic depth images from CT scans, more works are investigating the generation of point clouds from CT scans.
\citet{Chougule2013ConversionsOC} present a slice-based approach using automatic thresholding and edge detection to generate point clouds from CT scans.  
A voxel-based approach to generate surfaces from medical 3D data is the Marching Cubes Algorithm (MCA) \citep{lorensenMarchingcubes}.
\citet{SaitiMultimodalregistration} use the MCA to create synthetic point clouds from CT scans to learn multimodal registration with point clouds and CT scans.

\citet{RybakovSimultaneousGeneration} present a tool for generating pairs of radiographs and depth images from the XCAT phantom \citep{SegarsXCAT} under motion to support simulation based research on motion-related imaging scenarios.
In contrast, our work focuses on a clinically realistic data-generation framework that generates pairs of radiographs and depth images, based on CT scans of real patients and multiple realistic pose variations and acquisition parameters, in order to create realistic synthetic training data for learning-based patient pose assessment and to improve transfer to real datasets.
To the best of our knowledge, there is no existing framework with the same objectives and intended application. 


\section{Framework}
\label{framework}
\begin{figure*}[tbp]
    \centering
    \includegraphics[width=0.9\linewidth]{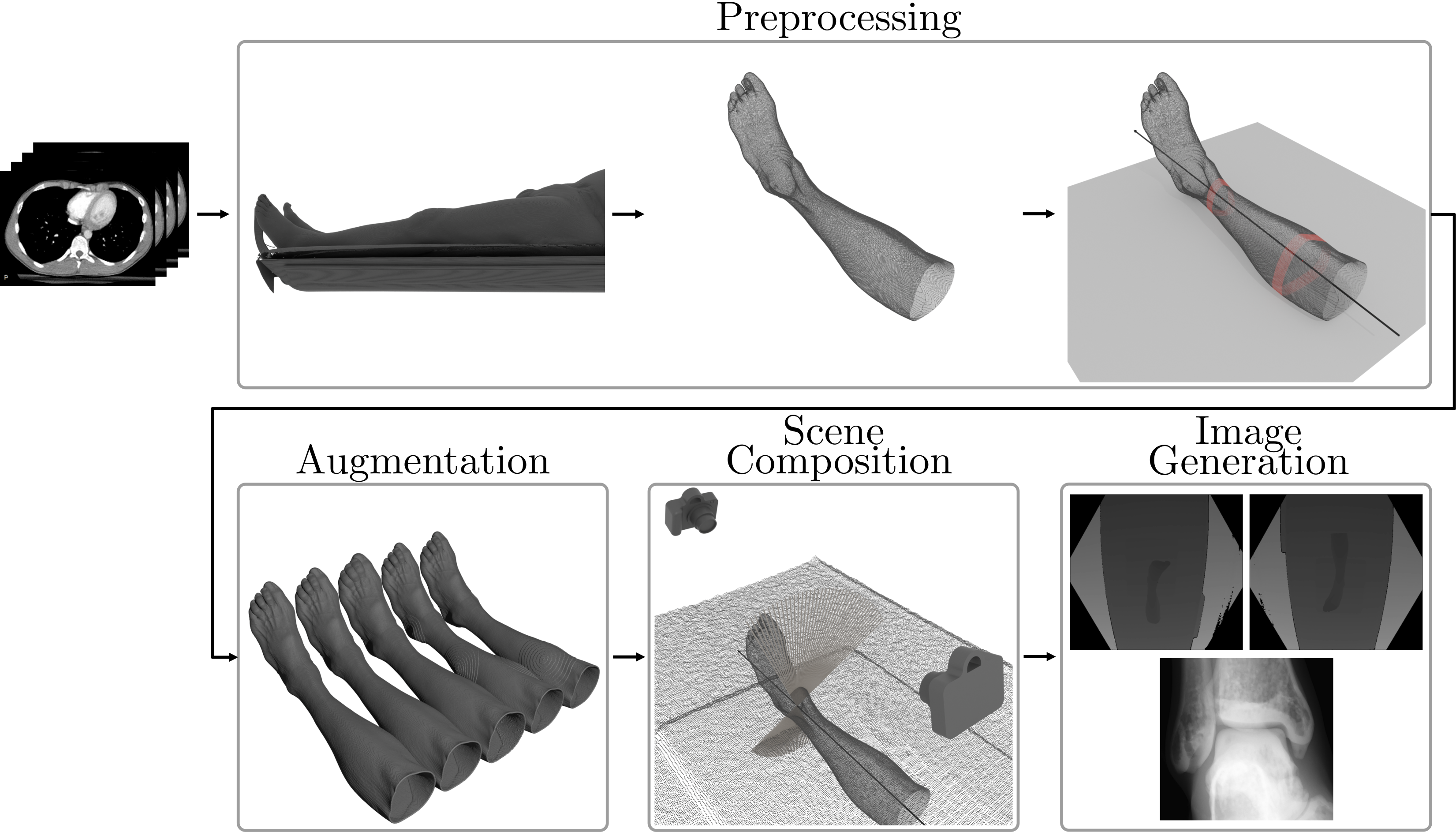}
    \caption{Schematic overview of the proposed framework.
    The CT scan as input is first transformed into a point cloud of the body surface, then the target anatomy is extracted and a rotation axis as well as a plane are defined.
    After augmentations of the point cloud, the target anatomy can be placed on the recorded patient table of the X-ray room so that realistic depth images can then be generated of different poses from different camera perspectives along with the corresponding radiographs. 
    }
    \label{fig:framework-schema}
\end{figure*}

The generation of synthetic radiographs and depth images from CT scans involves several steps. First, the surface of the target anatomy is extracted from the CT scan as a point cloud.
The point cloud is then augmented to simulate different body types.
These point clouds are placed on a pre-recorded point cloud of an imaging table in an X-ray room and rotated to create poses of different diagnostic quality. A synthetic depth image is generated for each pose by 2D projections of the point clouds.
The corresponding synthetic radiograph is generated for each pose from the CT scan using a MC simulation.
Our framework, which is implemented via Open3D's \citep{zhouOpen3D} graphical visualization, is shown in \Cref{fig:framework-schema} and examples of synthetically generated depth images and radiographs are shown in \Cref{fig:synthetic-images} in \Cref{example-images}. 
All these steps within the framework can be performed and modified semi-automatically through user input, resulting in a user-friendly and efficient workflow.
Although full automation within the framework may appear advantageous from a scalability perspective, and while we believe that a fully automated data generation is possible in principle, we consider human oversight essential in this context to ensure data quality.

The individual steps are described in more detail in the following.

\subsection{Preprocessing}
\label{framework-preprocessing}
In order to generate a synthetic depth image of the target anatomy from a CT scan, the target anatomy must first be extracted from the CT.
For this, the scan is converted to a point cloud using MCA.
The threshold value of the MCA is set to -500 Hounsfield units (HU) so that the air around the patient is removed.
This conversion from a CT array to a point cloud $P_{pat}$ includes a transformation $T^P_{CT}$ from the CT coordinate system $CT$ to the patient coordinate system $P$.
The point cloud is then cropped to the target anatomy to simplify the subsequent steps and calculations.
Although this step may cause the target anatomy to appear artificially truncated in the generated depth images, subsequently defined ROIs ensure that only the anatomically relevant regions are used as model input.
See \Cref{fig:synthetic-images} in \Cref{example-images} for examples.
Since only the surface of the target anatomy is relevant for the synthetic depth image, additional MCA runs, the clustering algorithm DBSCAN \citep{esterDBSCAN}, and cropping are applied to remove the imaging table and points that do not belong to the surface of the target anatomy.
In order to generate poses with different diagnostic quality, including inadequate quality, the target anatomy can be brought into other poses by specific rotations of the point cloud.
The axis of rotation is strongly dependent on the target anatomy.
For the upper ankle joint, it is the longitudinal axis, which is positioned in the point cloud such as to mimic human leg rotations.
The axis can be determined manually by computing the centers of two areas of the point cloud of the target anatomy. 
By selecting areas proximal to the ankle joint and areas around the calf, the axis passes directly through the center of the ankle joint, allowing for a realistic simulation of a patient's rotation. 
Finally, the preprocessing step can be completed by placing a plane $\Pi_P$ directly beneath the target anatomy to simplify further calculation steps.
For the upper ankle joint, the selection of three points from the point cloud is recommended for defining the plane equation: one in the distal area and two in the proximal area. The above described selection of points can be adjusted manually.
See \Cref{fig:rotaion-axis} in \Cref{framework-renderings} for illustrations.

As stated, these preprocessing steps are partly anatomy-specific, in particular the definition of the rotation axis and the plane $\Pi_P$. Accordingly, transferring the framework to other target anatomies may require corresponding setup adaptations. However, the overall generation pipeline is not inherently limited to the upper ankle joint and can in principle be adapted to other joints or limbs.

\subsection{Augmentation}
\label{augmentation}
By determining the normal vectors of the point cloud $P_{pat}$, it is possible to shift the point cloud both outwards and inwards along the direction of the normal vector, resulting in two additional point clouds.
This allows us to simulate patients with different shapes and sizes, and to increase the amount of data.
In order to further simulate patients with swellings or edema, it is possible to additionally shift points in certain areas along the normal vector by using a Gaussian distribution to simulate bumps.
The ankle joint itself and the tibia are particularly suitable for placing such bumps, since injuries frequently occur in these areas.
To simulate multiple and diverse injuries, we added 2-3 bumps locally within two additional point clouds, resulting in a total of four augmented point clouds.
The shift along the vector, as well as the position of the bumps and their size, can be adjusted manually to ensure realistic augmentations; see \Cref{fig:pc-augmentation} in \Cref{framework-renderings}.
The diagnostic quality later labeled on the corresponding synthetic radiographs applies equally to all augmentations of a particular pose, as it can be assumed that the anatomy that determines the quality, in particular the position of the bones in relation to each other, does not change with minor displacements along the normal direction.

\subsection{Scene Composition and Synthetic Depth Image Generation}
To generate realistic synthetic depth images, it is beneficial to embed the target anatomy in a realistic scene.
This can be achieved by recording the X-ray room in advance, including the imaging table, so that the target anatomy can then be placed on the imaging table under the X-ray device; see \Cref{fig:xray-placement} in \Cref{framework-renderings}.
The first step for this is to select the range of rotation of the target anatomy in order to include further non-diagnostic poses.
This rotation is described by the transformation $T_{rot}$.
Afterwards, the rotated target anatomy can be placed on the patient table.
The exact position of the target anatomy is selected so that the X-ray beam of the X-ray device passes through the axis of rotation of the target anatomy.
The placing on the table consists of a matching step and an optimization step, which are described by transformations $T_{mat}$ and $T_{opt}$, respectively.
For the matching step, the plane equation of the patient table $\Pi_T$ is first determined.
Note that the assumption that a plane exists for every X-ray examination, even if the patient is not radiographed on a table, is justified since the detector always defines a plane that can be matched. 
Now the plane $\Pi_P$ introduced in \Cref{framework-preprocessing} can be matched with $\Pi_T$ resulting in the transformation $T_{mat}$.
However, since $\Pi_P$ was only determined for the patient's original pose in the CT scan, it is possible that after rotation $T_{rot}$, the target anatomy is not realistically positioned on the table, and parts of the point cloud are located below the table, for example, see \Cref{fig:misalignment-pc-plane} in \Cref{misalignment-pc-plane}.
In order to avoid introducing bias into the generated data, the optimization step is necessary to ensure that the target anatomy is always placed realistically on the table. 
For the upper ankle joint, such optimization is possible by connecting the proximal and distal points of $P_{pat}$ that are closest to $\Pi_T$ as a line, aligning this line parallel to $\Pi_T$, and then translating it to $\Pi_T$.
\Cref{alg:closest-points} as well as \Cref{alg:optimization-algo} describe the optimization algorithm in more detail.
The rotation $T_{rot}$ and the placing of the target anatomy in the X-ray room using $T_{mat}$ and $T_{opt}$ can be described as a further transformation $T^{ToF}_{P}$ from $P$ to the ToF coordinate system $ToF$.
From the point clouds of the target anatomy and the X-ray room, a 2D projection yields the final synthetic depth image by taking into account intrinsic and extrinsic camera parameters and distortion coefficients.
This way, synthetic depth images can be generated for any desired angles of rotation, augmentations, and camera views.
The resulting flexibility of the framework has already been used to investigate suitable camera positions for pose assessment in \cite{lauferEvaluationofToFpositioning}. Furthermore, because the scene composition explicitly incorporates the recorded room geometry together with the intrinsic and extrinsic camera parameters, the framework is not inherently restricted to the specific X-ray room and camera configuration used in this study. In principle, it can be adapted to different X-ray rooms, X-ray devices, clinical environments, and camera configurations by replacing the corresponding scene and acquisition parameters, thus allowing the generation of custom synthetic datasets for specific acquisition settings.

  
  
  
  
  

\begin{algorithm}
    \caption{ClosestPoints}
    \label{alg:closest-points}

    \begin{algorithmic}[1]
      \Require $P - Point\ cloud$, $\Pi - Plane$
      \Ensure $p_{front}, p_{back}, d_{front}, d_{back}$
      \Statex \textit{\# split $P$ into front and back subsets }
      \State $(P_{front}, P_{back}) \gets \textsc{SplitFrontBack}(P)$
      \Statex  
    
      \Statex \textit{\# closest points to $\Pi$ in each subset}
      \State $p_{front} \gets \textsc{ClosestPointToPlane}(P_{front}, \Pi)$
      \State $p_{back}  \gets \textsc{ClosestPointToPlane}(P_{back}, \Pi)$
      \Statex  
    
      \Statex \textit{\# corresponding minimal distances}
      \State $d_{front} \gets \textsc{PlaneDistance}(p_{front}, \Pi)$
      \State $d_{back}  \gets \textsc{PlaneDistance}(p_{back}, \Pi)$
      \Statex  
    
      \State \Return $(p_{front}, p_{back}, d_{front}, d_{back})$
    \end{algorithmic}
\end{algorithm}

\begin{algorithm}
    \caption{Optimization algorithm to find $T_{opt}$}
    \label{alg:optimization-algo}
    \begin{algorithmic}[1]
      \Require $P_{pat} - Patient\ point\ cloud$, $\Pi_T - Table\ plane$, $\tau - Threshold$
      \Ensure $T_{opt} - Optimized\ transformation$

      \State $R_{opt} \gets I_4$
      \Statex  
    
      \Statex \textit{\# initial closest points to $\Pi_T$ and distances}
      \State $(p_1, p_2, d_1, d_2) \gets \textsc{ClosestPoints}(P_{pat}, \Pi_T)$
      \Statex  
    
      \While{$|d_1 - d_2| > \tau$}
        \Statex \textit{\# angle $\alpha$ so line$(p_1,p_2)$ becomes parallel to $\Pi_T$ }
        \State $\alpha \gets \textsc{LinePlaneAlignment}(\Pi_T, \text{line}(p_1, p_2))$
        \Statex  
    
        \Statex \textit{\# rotate with $R_X(\alpha)$ around X axis and update}
        \State $R \gets R_X(\alpha)$
        \State $P_{pat} \gets R \cdot P_{pat}$
        \State $R_{opt} \gets R \cdot R_{opt}$
        \Statex  

        \Statex \textit{\# recompute closest points}
        \State $(p_1, p_2, d_1, d_2) \gets \textsc{ClosestPoints}(P_{pat}, \Pi_T)$
      \EndWhile
      \Statex 
    
      \Statex \textit{\# final translation onto $\Pi_T$}
      \State $d \gets \max(d_1, d_2)$
      \State $T_{plane} \gets \textsc{TranslationOntoPlane}(\Pi_T, d)$
      
      \State $P_{pat} \gets T_{plane} \cdot P_{pat}$
      \Statex 
    
      \Statex \textit{\# combined optimized transformation}
      \State $T_{opt} \gets T_{plane} \cdot R_{opt}$
      \Statex 
    
      \State \Return $T_{opt}$
    \end{algorithmic}
    \end{algorithm}

\subsection{Synthetic Radiograph Generation}
When generating synthetic radiographs, it is important that all depicted anatomical features are identical to those of real radiographs in the same position to avoid false labels of the corresponding depth images.
Generation based on a physical model was therefore preferred to methods based on deep learning.
Since only a region smaller than the one depicted in the point clouds described in \Cref{framework-preprocessing} must be visible in the radiograph, the target anatomy is cropped out of the CT in the first step.
The cropped CT voxels are then converted into material and mass-density voxels. 
The material voxels each contain one of the materials: air, soft tissue, bone, or titanium, based on the HU value of the respective voxel.
Similarly, the mass-density voxels contain the density of the material adjusted by the HU value.

For each pair of material and mass-density voxels, multiple radiographs corresponding to different positions are generated.
Instead of changing the position of the target anatomy, which would require rotation and interpolation of the voxels, the position of the X-ray device relative to the target anatomy is changed; see \Cref{fig:xray-pc} in \Cref{framework-renderings}. 
The corresponding position of the X-ray device in the CT coordinate system can be obtained using the inverted transformations $(T^{ToF}_P)^{-1}(T^P_{CT})^{-1}$.
To generate the radiograph, the MCGPU tool \citep{BADAL2011813} was used for a MC simulation together with the material properties from the PENELOPE 2006 material files \citep{salvat2006penelope} to simulate \(2\cdot10^{10}\) X-ray beam paths.
While more simulated paths would reduce the noise in the generated radiographs, the simulation time would increase. 
The use of \(2\cdot10^{10}\) simulated paths allowed us to create realistic radiographs in a reasonable amount of time.
The resulting raw image is then converted to a synthetic radiograph using a non-linear value mapping, to obtain a look similar to a real radiograph.
The synthetic radiograph is not as detailed as a real radiograph, but expert radiologists have validated that the visual quality is suitable for assessing the diagnostic quality for a given pose.

\section{Datasets}
\label{datasets}
The three datasets used in this paper all consist of depth images from two camera views.
While the real-world clinical dataset consists exclusively of depth images, the synthetic dataset and the anatomical preparations dataset also contain the corresponding radiographs.
Each of these radiographs was assessed by 4 radiologists to determine its diagnostic quality on a scale of 1 to 3 in steps of 0.5, making 0.5 the smallest increment of the annotation scale.
A diagnostic quality of 1 is ideal and a diagnostic quality of 3 is inadequate.
The deciding factor in the assessment of the upper ankle joint is the visibility of the joint space; see \Cref{fig:synthetic-images} in \Cref{example-images}. 
Note that correct visualization of joint spaces requires proper rotational alignment of the involved bones with respect to the beam direction, which is typically achieved by positioning the patient based on external anatomical landmarks.
Because even slight rotations are sufficient to partially obscure the joint space, and because the osseous structures that ultimately determine whether the joint space is correctly visualized are not directly visible during positioning, the resulting diagnostic quality is highly dependent on the radiographers performing the positioning.
Accordingly, radiographs with labels in the interval of [1, 2.5) can be classified as \textit{diagnostic}, whereas radiographs with labels greater than or equal to 2.5 are considered \textit{non-diagnostic} and should be repeated in practice.
This threshold therefore follows directly from the radiologists' annotation scheme and their clinical assessment of diagnostic usability of a radiograph.
Note that, to further increase the scalability of the framework, the synthetic radiographs could also be labeled automatically based on their quality, for example using the framework presented in \citet{mairhoferAIbasedFrameworkDiagnostic2021}. However, we still consider expert manual assessment essential to ensure data quality.

\subsection{Synthetic Dataset}
\label{synthetic-dataset}
Using the framework proposed in \Cref{framework}, we were able to generate pairs of synthetic radiographs and depth images of upper ankle joints from 10 CTs of different patients in 3077 different poses. 
The anonymized CT scans were selected to contain flexed upper ankle positions and exclude clutter such as tubes or screws.
From the 10 CTs, a total of 17 upper ankle joints were extracted and rotated medially around the longitudinal axis in a range of 90 degrees.
A synthetic depth image was generated from two camera views $V_1$ and $V_2$ for each half degree, i.e. a total of 181 poses per foot. 
This was done for the original point cloud and the four augmented variants, resulting in a total of 30770 depth images. 
Since the synthetic radiograph is the same for all camera views and augmentations, one synthetic radiograph was created for each of the 181 poses resulting in a total of 3077 synthetic radiographs. 
To the best of our knowledge, this is by far the largest dataset linking depth images to diagnostic quality.

\subsection{Anatomical Preparations Dataset}
\label{anatomic-preparation-dataset}
As presented in \citet{lauferPatientPoseAssessment2024}, we captured two anatomical preparations \textendash{} a left and a right lower leg of two women in 174 different poses using two Microsoft Azure Kinect ToF cameras.
These cameras were attached to a Philips DigitalDiagnost C90 using a custom-made 3D-printed mount.
Parallel to the depth images from two different views, a radiograph of the upper ankle joint was also taken.
The preparations were not only rotated medially around the longitudinal axis but also flexed in three different positions of the ankle joint.
More detailed information on this published dataset can be found in \citet{lauferPatientPoseAssessment2024}.

\subsection{Real-World Clinical Dataset}
\label{real-world-clinical-dataset}
Although with the anatomical preparation dataset we have acquired a dataset that contains real depth images and radiographs of real target anatomies, it only includes data from two subjects. 
We have therefore decided to acquire another dataset with additional living subjects in a scenario that mirrors clinical practice.
The dataset was collected in an X-ray room used in clinical practice, and the camera-X-ray device configuration was identical to that used in the anatomical preparations dataset.
Ten male and eight female subjects aged between 21 and 33 were positioned in three realistic positions per foot for an X-ray examination of the upper ankle joint. 
The three positions were identical in terms of flexion, but differed in terms of rotation around the longitudinal axis \textendash{} one pose was the textbook position for optimal positioning, and two poses diverged from this position. 
As a result, we were able to capture depth images of 108 different poses.
As none of the subjects had any indications, we were not permitted to take the corresponding radiographs for labeling the diagnostic quality of the pose.
The poses can therefore only be categorized as good or bad based on the positioning. For simplicity, these categories were assigned labels of 1 and 3 for training. However, these labels are not radiographic ground truth, but surrogate labels, which makes this a weakly labeled dataset.
Given the geometry of the foot and the narrow rotational range within which a diagnostic quality of 1 is even achievable (see \Cref{fig:synthetic-images} in \Cref{example-images}), it is highly unlikely that a poor pose would yield a better diagnostic quality than the textbook pose, or even a diagnostic quality of 1. 
However, this does not guarantee that the textbook pose always corresponds to a true diagnostic quality label of 1.
Nevertheless, in addition to increasing the number of subjects and including living participants, the purpose of this dataset is to evaluate whether poses can be assessed and ranked consistently across multiple subjects in a clinically realistic setting.
\begin{figure*}[htb]
    %
    {
        \subfigure{%
            \label{fig:sep-network}
            \includegraphics[width=0.32\linewidth]{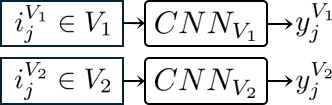}
        }
        \subfigure{%
            \label{fig:camera-specific-view}
            \includegraphics[width=0.319\linewidth]{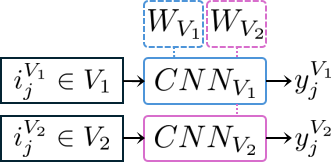}
        }
        \subfigure{%
            \label{fig:unified-camera-view}
            \includegraphics[width=0.32\linewidth]{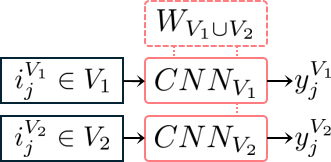}
        }
        \caption{
           \ref{fig:sep-network} shows the separate network architecture: individual images $i^{V_1}_j$ and $i^{V_2}_j$ from each camera view $V_1$ and $V_2$ are used to train two separate CNNs.
           $y^{V_1}_j$ and $y^{V_2}_j$ are the continuous outputs of the networks, ranging from 1 to 3. 
           \ref{fig:camera-specific-view} shows the \textit{camera view specific approach}, where the CNNs are initialized with the weights $W_{V_1}$ and $W_{V_2}$ obtained by pretraining with the corresponding view.
           $CNN_{V_1}$ would therefore only be initialized with weights $W_{V_1}$ that were obtained by pretraining with images only from $V_1$.
           \ref{fig:unified-camera-view} shows the \textit{unified camera view approach}, where both CNNs are initialized with the same weights $W_{V_1 \cup V_2}$ obtained by pretraining with images from both views $V_1 \cup V_2$.
        }
    }
\end{figure*}
The clinical routine was replicated during the acquisition process. 
This means that the X-ray device and the cameras attached to it were often moved to accommodate the specific patient before positioning. 
This change in the camera-target anatomy relation, as well as the position of the detector under the target anatomy, adds to the diversity of the dataset in addition to the increased number of subjects. 
Furthermore, instead of capturing a single depth image per chosen pose, a time interval was captured, so that small movements of the target anatomy, characteristic camera noise, and invalid points are included.
This also corresponds to a potential real-world implementation, so that in general, the real-world clinical dataset differs greatly from the synthetic dataset and the anatomical preparations dataset, making generalization challenging.
Examples of the dataset are shown in \Cref{fig:real-world-images} in \Cref{real-world-examples}.


\section{Experiments and Training}
\label{experiments-and-training}

The two experiments carried out are designed to answer the following research questions:
\begin{enumerate}
    \item[] \textbf{Question 1}: Can a neural network be trained on the generated synthetic depth images to assess the pose with high accuracy?
    \item[] \textbf{Question 2}: If so, how useful are the features learned from the synthetic dataset for real-data pose assessment, both in terms of downstream finetuning and linear probing?
\end{enumerate}

\begin{figure*}[!h]
    \centering
    \includegraphics[width=0.7\linewidth]{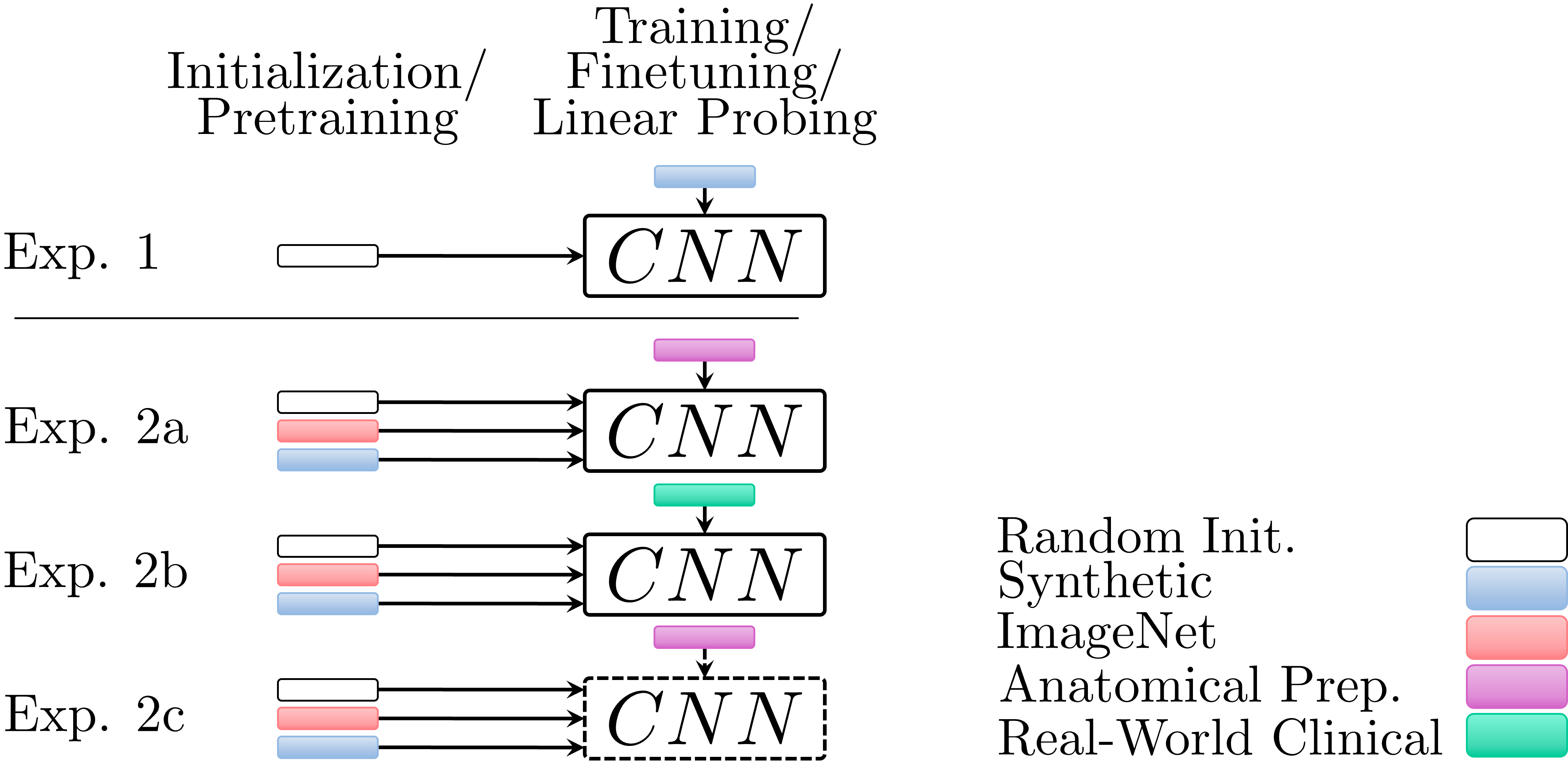}
    \caption{
    Schematic overview of the experiments conducted. The colored boxes to the left of the CNN model represent the datasets used for pretraining the model, or for random initialization. The colored boxes above the CNN model represent the datasets used for training, finetuning, linear probing, and the corresponding testing. The solid lines of the model indicate training/finetuning, and the dashed lines indicate linear probing. Note that this is a simplified representation and that, for example, pretraining with the synthetic dataset, as described in \Cref{pretraining}, can be further subdivided.}
    \label{fig:exp-overview}
\end{figure*}

\subsection{Experiment 1}
\label{experiment-1}
The first experiment aims to address the first research question by evaluating whether relevant features can be learned solely through the use of the synthetic dataset.
To this end, a randomly initialized model was trained and tested on the synthetic dataset using three-fold cross-validation (see Exp. 1 in \Cref{fig:exp-overview}).
For each cross-validation run the test set consisted of three randomly selected upper ankle joints so that no subject from the test set would appear in the training set.

In addition, to clarify whether the augmentation of the depth images (see \Cref{augmentation}) has a benefit, the training was carried out both without augmentation and with augmentation, and separately for the different augmentation types size, bumps, and both size and bumps (all).
Note that possible differences in the results in comparison to \citet{lauferPatientPoseAssessment2024} are due to different implementations and updated libraries.
The results are shown in \Cref{table:synthetic-dataset}.

 \begin{table*}[htb]
    \centering
    \begin{minipage}[h]{\linewidth}
        \centering
        \caption{Results of the experiments conducted solely on the synthetic dataset and evaluated by the metrics described in \Cref{metrics} regarding the impact of different augmentation of the synthetic depth images. We distinguish between training without augmentation, size augmentation, bump augmentation, and all augmentations combined; see \Cref{augmentation}.
        Note that training with augmented data can generally improve the results for all metrics.}
    \label{table:synthetic-dataset}

        {
  \begin{tabular}{lc|c|c|c}

    \toprule
    Metric 
        & \multicolumn{1}{c}{without aug.} 
        & \multicolumn{1}{c}{size aug.}
        & \multicolumn{1}{c}{\makecell{bump aug.}}
        & \multicolumn{1}{c}{\makecell{all aug.}} \\
    \midrule
    MAE 
        & 0.23{\scriptsize$\pm$0.02} 
        & \textbf{0.21}{\scriptsize$\pm$0.02}
        & 0.23{\scriptsize$\pm$0.02}
        & 0.22{\scriptsize$\pm$0.02} \\
    Correlation $r_s$ 
        & 0.85{\scriptsize$\pm$0.03} 
        & \textbf{0.87}{\scriptsize$\pm$0.02}
        & 0.85{\scriptsize$\pm$0.03}
        & 0.86{\scriptsize$\pm$0.03} \\
    \midrule
    Accuracy [\%] 
        & 85.29{\scriptsize$\pm$2.34} 
        & \textbf{87.6}{\scriptsize$\pm$2.6}
        & 85.88{\scriptsize$\pm$2.59}
        & 86.9{\scriptsize$\pm$2.56} \\
    Diag. Acc. [\%] 
        & 85.45{\scriptsize$\pm$2.77} 
        & \textbf{86.96}{\scriptsize$\pm$2.47}
        & 85.17{\scriptsize$\pm$2.78}
        & 86.94{\scriptsize$\pm$2.84} \\
    \hspace{0.25em} Sens.\hspace{0.29em}[\%] 
        & 91.5{\scriptsize$\pm$1.5} 
        & \textbf{92.82}{\scriptsize$\pm$1.8}
        & 92.02{\scriptsize$\pm$1.64}
        & 91.9{\scriptsize$\pm$2.22} \\
    \hspace{0.25em} Spec.\hspace{0.29em}[\%]
        & 76.9{\scriptsize$\pm$6.01} 
        & 79.0{\scriptsize$\pm$5.88}
        & 75.72{\scriptsize$\pm$6.37}
        & \textbf{80.21}{\scriptsize$\pm$5.29} \\
    \bottomrule
\end{tabular}
}
\end{minipage}
\end{table*}

\subsection{Experiment 2}
\label{experiment-2}
To answer the second question, we evaluated whether pretraining with the synthetic data improves performance when finetuning on real data, in our case on the anatomical preparations dataset and the real-world clinical dataset (see \Cref{anatomic-preparation-dataset} and \Cref{real-world-clinical-dataset}).
The experiments conducted for this purpose can be divided into three parts.

In Experiment 2a we compare models with randomly initialized weights, i.e. trained from scratch, with models pretrained on ImageNet  \citep{dengImageNetLargescaleHierarchical2009} and on the synthetic dataset, all of which are subsequently finetuned on the anatomical preparations dataset (see Exp. 2a in \Cref{fig:exp-overview}).
Furthermore, as in \Cref{experiment-1}, we evaluated whether the augmentation of the depth data during pretraining has an influence on the results.
In addition, we considered both unified and camera view specific pretraining strategies, which are described in more detail in \Cref{pretraining}. 
As the anatomical preparations dataset consists of only two preparations, for finetuning we trained on one preparation and tested on the other, and vice versa.
The results of Exp. 2a are shown in \Cref{table:anatomical-preparation}.

With Experiment 2b we then investigate whether the findings from these experiments can be generalized to the finetuning with the real-world clinical dataset.
For this, we compare models trained from scratch on the real-world clinical dataset with models pretrained on ImageNet and with camera view specific pretraining on the synthetic dataset using selected augmentation settings, both followed by finetuning on the real-world clinical dataset (see Exp. 2b in \Cref{fig:exp-overview}).
For the real-world clinical dataset, we used subject-level three-fold cross-validation for training and testing.
The results are shown in \Cref{table:real-world}.

To finally evaluate how useful the features learned during pretraining actually are, using linear probing in Experiment 2c we compare the performance on the anatomical preparations dataset of a model that was trained from scratch, a model pretrained on ImageNet, and models pretrained on the synthetic dataset using both unified and view-specific pretraining (see Exp 2c in \Cref{fig:exp-overview}).
The training/test split for the anatomical preparation dataset is identical to Experiment 2a and the results are shown in \Cref{table:linear-probing}.

\subsection{Training}
\label{training}

\paragraph{Implementation Details}
All experiments are modeled as regressions, to better reflect the ordered nature of the labels.
While the depth images served as input for the models, the models output the diagnostic quality as a single continuous value between 1 and 3.
All experiments were implemented using PyTorch and repeated 10 times with different seeds. 
The results are shown as average and standard deviation.

As models, two EfficientNet-B0 \citep{tanEfficientNetRethinkingModel2019}, called \(CNN_{V_1}\) and \(CNN_{V_2}\) were used.
We deliberately chose this comparatively small but powerful model with rather few parameters.
Besides being better matched to the relatively small size of our realistic datasets, this choice was primarily motivated by practical deployment requirements: 
If necessary the model should be able to run on a CPU to keep cost and complexity as low as possible, while also being fast enough to process frames in real-time from the live feed of a ToF camera to provide immediate feedback to the radiographer.
For training, we followed the separate network architecture (see \Cref{fig:sep-network}). As described in \citet{lauferPatientPoseAssessment2024}, the \(CNN_{V_1}\) is only trained on the depth images from camera view \(V_1\) while the \(CNN_{V_2}\) is trained on images from camera view \(V_2\).
The models each receive a single channel depth image \(i\), resized to \(336 \times 336\) pixels, as input and output a single continuous value \(y\).
The models are trained to minimize the Mean Squared Error (MSE; see \Cref{mse-def} in \Cref{metrics-definition}) between the model's output \(y\) and the averaged diagnostic quality \(x\) that had been assigned to the input depth image.
The Adam optimizer \citep{kingmaAdamMethodStochastic2017} was used with an initial learning rate of $10^{-3}$, which was decreased 4 times by a factor of 10 in the last 10,000 steps.
We used a batch size of 16 and trained each model for 200,000 steps.

\paragraph{Pretraining}
\label{pretraining}
There are four alternative ways of pretraining to initialize weights before finetuning.
For training \textit{from scratch}, without any pretraining, the models are randomly initialized.
For the \textit{pretraining on ImageNet}, both models \(CNN_{V_1}\) and \(CNN_{V_2}\) are initialized with the ImageNet weights.
When pretraining using the synthetic dataset there are two possible ways, as this dataset also contains depth images of two views:
In the \textit{unified camera view approach}, both models are initialized with identical weights \(W_{V_1 \cup V_2}\) obtained by pretraining with images from both camera views \(V_1\) and \(V_2\) of the synthetic dataset; see \Cref{fig:unified-camera-view}.
For the \textit{camera view specific approach}, each model \(CNN_{V_1}\) and \(CNN_{V_2}\) is initialized with weights obtained by pretraining on synthetic depth images from only camera view \(V_1\) or \(V_2\) respectively; see \Cref{fig:camera-specific-view}. 
This approach effectively halves the amount of pretraining data, but the finetuning is more specific.
The pretraining on the synthetic dataset was performed as described above in \Cref{experiment-1}, but for the fact that the whole dataset was used as training set. 

\paragraph{Finetuning}
Using the separate network architecture (see \Cref{fig:sep-network}), the training from scratch and finetuning on the anatomical preparations dataset were performed with the same hyperparameters as listed above in order to be comparable.
Since the radiographs were not available in the real-world clinical dataset, this can also be described as a weakly labeled dataset.
In order to ensure comparability with the other experiments, we assigned label 1 to the deliberately well-positioned poses and label 3 to the poor ones for training and testing.
Since we did not just capture one depth image per pose in this experiment, but rather a time series, we sampled 5 consecutive frames for each pose for both training and testing.

\paragraph{Linear Probing}
Even though only the final regression head is trained in this experiment and the previous backbone remains frozen during training, we conducted the experiments with the same hyperparameters as used for finetuning and the same number of training steps for comparability reasons. 

Across all experiments, we conducted 21 experimental configurations.
Counting the different pretraining variants, the two separately trained view-specific networks, the different training/test splits, and the 10 repetitions with different random seeds, this amounts to 980 downstream training runs and 1100 trained models when the synthetic pretraining runs are included.

\begin{table*}[htb]
    \centering
    \hfill

    \scriptsize
    \setlength{\tabcolsep}{2.9pt}
    \begin{minipage}[t]{\linewidth}
         \caption{Results of the experiments without pretraining, with pretraining on ImageNet and pretraining on the synthetic dataset and subsequent finetuning on the anatomical preparations dataset, using the metrics and methods described in \Cref{experiments-and-training}.
         Note that pretraining with synthetic data outperforms models without pretraining and with pretraining on ImageNet.}
         \label{table:anatomical-preparation}
        
        {
    \begin{tabular}{@{}>{\small}l>{\scriptsize}r|>{\scriptsize}r|
                *{4}{>{\scriptsize}c}|*{4}{>{\scriptsize}c}@{}}
      \toprule
      \multirow{3}{*}{\makecell{Metric}}       
          & \multirow{3}{*}{\makecell{\small from\\\small scratch}} 
          & \multirow{3}{*}{\makecell{\small pretrained \\\small on \\\small ImageNet}}
          & \multicolumn{8}{c}{\small pretrained on synthetic dataset} \\
          &                                &                                      
          & \multicolumn{4}{c|}{\small unified camera view} 
          & \multicolumn{4}{c}{\small camera view specific} \\
          &                                &                                      
          & \small wo/ aug. 
          & \small size aug. 
          & \small bump aug. 
          & \small all aug. 
          & \small wo/ aug. 
          & \small size aug. 
          & \small bump aug.
          & \small all aug. \\ 
        
      \midrule
      MAE                                      
          & 0.28{\scriptsize$\pm$0.04}     
          & 0.31{\scriptsize$\pm$0.05}           
          & \textbf{0.22}{\scriptsize$\pm$0.04}  
          & 0.24{\scriptsize$\pm$0.04}           
          & 0.28{\scriptsize$\pm$0.06}           
          & 0.24{\scriptsize$\pm$0.04}           
          & 0.23{\scriptsize$\pm$0.03}           
          & \textbf{0.22}{\scriptsize$\pm$0.06}  
          & 0.24{\scriptsize$\pm$0.04}           
          & 0.24{\scriptsize$\pm$0.03}           \\
      Cor. \(r_s\)                      
          & 0.88{\scriptsize$\pm$0.04}     
          & 0.9{\scriptsize$\pm$0.03}            
          & \textbf{0.92}{\scriptsize$\pm$0.03}  
          & 0.9{\scriptsize$\pm$0.04}            
          & 0.88{\scriptsize$\pm$0.05}           
          & 0.91{\scriptsize$\pm$0.03}           
          & 0.9{\scriptsize$\pm$0.03}            
          & 0.91{\scriptsize$\pm$0.04}           
          & 0.89{\scriptsize$\pm$0.04}           
          & 0.89{\scriptsize$\pm$0.03}           \\ 
      \midrule
      Acc. \hspace{0.29em}[\%]             
          & 79.25{\scriptsize$\pm$6.98}    
          & 77.37{\scriptsize$\pm$8.22}          
          & 89.03{\scriptsize$\pm$5.51}          
          & 86.91{\scriptsize$\pm$6.14}          
          & 82.16{\scriptsize$\pm$8.67}          
          & 86.85{\scriptsize$\pm$6.05}          
          & \textbf{90.45}{\scriptsize$\pm$4.73} 
          & 90.07{\scriptsize$\pm$7.64}          
          & 86.69{\scriptsize$\pm$6.6}           
          & 88.92{\scriptsize$\pm$4.35}          \\
      Diag. Acc.\hspace{0.29em}[\%]            
          & 89.08{\scriptsize$\pm$3.75}    
          & 84.2{\scriptsize$\pm$2.43}           
          & 90.93{\scriptsize$\pm$3.0}           
          & 91.91{\scriptsize$\pm$4.64}          
          & 88.83{\scriptsize$\pm$5.77}          
          & 90.49{\scriptsize$\pm$5.09}          
          & 92.8{\scriptsize$\pm$2.14}  
          & 92.43{\scriptsize$\pm$4.05}          
          & \textbf{93.87}{\scriptsize$\pm$2.05}          
          & 93.51{\scriptsize$\pm$2.15}          \\
      \hspace{0.25em} Sens.\hspace{0.29em}[\%] 
          & 91.21{\scriptsize$\pm$4.23}    
          & 93.79{\scriptsize$\pm$3.19} 
          & 92.65{\scriptsize$\pm$7.45}          
          & 90.9{\scriptsize$\pm$10.29}          
          & \textbf{94.42}{\scriptsize$\pm$5.31}          
          & 93.31{\scriptsize$\pm$9.5}           
          & 87.64{\scriptsize$\pm$4.51}          
          & 91.16{\scriptsize$\pm$6.72}          
          & 92.08{\scriptsize$\pm$4.75}          
          & 91.24{\scriptsize$\pm$7.34}          \\
      \hspace{0.25em} Spec.\hspace{0.29em}[\%] 
          & 88.66{\scriptsize$\pm$4.61}    
          & 80.05{\scriptsize$\pm$4.6}           
          & 89.84{\scriptsize$\pm$5.49}          
          & 92.24{\scriptsize$\pm$6.59}          
          & 85.92{\scriptsize$\pm$10.43}         
          & 88.9{\scriptsize$\pm$8.1}            
          & \textbf{95.34}{\scriptsize$\pm$3.91} 
          & 93.1{\scriptsize$\pm$6.27}           
          & 94.88{\scriptsize$\pm$3.04}          
          & 94.62{\scriptsize$\pm$3.89}          \\ 
      \bottomrule
  \end{tabular}
  }
\end{minipage}
\end{table*}

\subsection{Metrics}\label{metrics}
When using two depth images $i^{V_1}_j$ and $i^{V_2}_j$ of the same pose from the two camera views as input, the continuous outputs $y^{V_1}_j$ and $y^{V_2}_j$  of the two models are averaged and compared with the corresponding quality label $x_j$. 
The following metrics are calculated:
\textbf{Mean Absolute Error} (MAE), \textbf{Spearman correlation coefficient} ($r_s$) (see \Cref{mae-def} and \Cref{rs-def} in \Cref{metrics-definition}), as well as two accuracies.
The \textbf{Accuracy} measures how often the prediction differs with less than 0.5 from the label:
\begin{equation*}
    Accuracy=\frac{1}{N} \sum_{j=1}^{N}\mathbbm{1}(|y_j - x_j| < 0.5), 
\end{equation*}
where \(y_j = (y^{V_1}_j + y^{V_2}_j) / 2\) is the average prediction, $N$ is the number of samples, $x_j$ is the label, and \(\mathbbm{1}\) is the indicator function.
The threshold of 0.5 follows directly from the annotation scale: since diagnostic quality was rated by the radiologists in increments of 0.5, a prediction is counted as accurate if it deviates by less than one annotation step from the assigned label.
The \textbf{Diagnostic Accuracy} measures how often the prediction of whether a depth image is diagnostic or non-diagnostic is correct, i.e., whether label and prediction are both below or above the 2.5 threshold:
\begin{align*}
    Diagnostic\ Accuracy 
  &= \frac{1}{N} \sum_{j=1}^{N} 
     \mathbbm{1}\Big( ({y}_j < 2.5 \land x_j < 2.5) \\
  &\qquad\lor ({y}_j \geq 2.5 \land x_j \geq 2.5) \Big)
\end{align*}
As described in \Cref{datasets}, this threshold follows directly from the radiologists' expert assessment, according to which labels below 2.5 correspond to diagnostically usable radiographs, whereas labels of 2.5 or 3 are considered non-diagnostic.
Therefore, the Diagnostic Accuracy is of particular importance, as it reduces the clinically critical decision of whether a radiograph must be retaken to a single interpretable value.
Since it is furthermore worse to classify an image that is not diagnostic as diagnostic than vice versa, the \textbf{Sensitivity} and \textbf{Specificity} are also calculated for the diagnostic accuracy.

Due to the weak labels of the real-world clinical dataset, the radiographic ground truth quality labels for the individual poses are not known and may potentially deviate from the assigned label of 1 and 3, especially when the subject moved during the five consecutive frames.
It is therefore useful to introduce additional metrics that do not evaluate the prediction with regard to these surrogate labels directly, but rather the ranking of the good pose in relation to the bad ones of the same foot.
For this reason, we introduce the \textbf{Pose Delta} metric, which calculates the difference between the averaged predicted quality of the bad poses and the averaged predicted quality of the good poses. The larger this value, the greater the average difference in the prediction of good and bad poses.
If we first define the network's average prediction per pose as follows:

\begin{equation*}
  \bar{y}_p = \frac{1}{T} \sum_{i=1}^{T} y_{p,i},
\end{equation*}
where $T$ is the number of frames for that pose and $y_{p,i}$ the prediction for frame $i$ of pose $p$, then the \textbf{Pose Delta} can be defined as:

\begin{equation*}
  Pose\ Delta
   = \frac{1}{N} \sum_{j=1}^{N}
    \Bigl(
      \frac{1}{2}(\bar{y}_{b_1,j} + \bar{y}_{b_2,j})
      - \bar{y}_{g,j}
    \Bigr),
\end{equation*}
where $N$ is the total number of subject-foot pairs, $\bar{y}_{g,j}$ is the averaged prediction of pair $j$ for the good pose, and $\bar{y}_{b_1,j}$, $\bar{y}_{b_2,j}$ are the averaged predictions of pair $j$ for the bad poses.
For the \textbf{Pose Margin Accuracy} metric, we count how often the difference between the predicted mean quality $\bar{y}$ of the best of the two bad poses ($\bar{y}_{b_1}$, $\bar{y}_{b_2}$) and the predicted mean quality of the good pose $\bar{y}_{g}$ of a subject-foot pair $j$ is greater than or equal to 0.5:
\begin{align*}
  Pose\ Margin\ Accuracy 
  = \frac{1}{N} \sum_{j=1}^{N}
    \mathbbm{1}\Big(
      \min\bigl(\bar{y}_{b_1,j}, \bar{y}_{b_2,j}\bigr)\\
      - \bar{y}_{g,j}
      \,\ge\, 0.5
    \Big).
\end{align*}
As in the Accuracy metric, the threshold of 0.5 corresponds to one annotation step on the original expert-defined quality scale. We chose this value both for consistency and because it represents the smallest non-trivial separation between good and bad poses.
Note that the reported metrics are used solely for evaluation, whereas model training is performed as a regression task using the MSE loss (see \Cref{training}).

\begin{table*}[htb]
    \centering
    \begin{minipage}[h]{\linewidth}
        \centering
        \caption{Results of the experiments without pretraining, with pretraining on ImageNet, and pretraining on the synthetic dataset and subsequent finetuning on the real-world clinical dataset, using the metrics and methods described in \Cref{experiments-and-training}.
        Since \emph{camera view specific} pretraining performed better when finetuning on the anatomical preparations dataset (see \Cref{table:anatomical-preparation}) on average, and size augmentation performed best when training with synthetic dataset (see \Cref{table:synthetic-dataset}), we also performed \emph{camera view specific} pretraining using size and all augmentations for comparison and finetuned on the real-world clinical dataset. 
        Note that with this realistic dataset, pretraining on the synthetic dataset also increases downstream performance.}
        \label{table:real-world}
        
        {
                      \begin{tabular}{@{}lr|r|cc@{}}
                \toprule
                \multirow{2}{*}{\makecell{Metric}}
                 & \multirow{2}{*}{\makecell{from                                             \\scratch}}
                 & \multirow{2}{*}{\makecell{pretrained                                     \\on ImageNet}}
                 & \multicolumn{2}{c}{pretrained on synthetic dataset}                        \\
                 &                   &                                  & size aug. & all aug. \\ 
                \midrule
                MAE                                      
                 & 0.39{\scriptsize$\pm$0.06}

                 & 0.38{\scriptsize$\pm$0.06}

                 & 0.35{\scriptsize$\pm$0.06}

                 & \textbf{0.32}{\scriptsize$\pm$0.06}                                        \\
                Correlation \(r_s\)                      
                 & 0.64{\scriptsize$\pm$0.07}

                 & 0.63{\scriptsize$\pm$0.06}

                 & 0.65{\scriptsize$\pm$0.06}

                 & \textbf{0.67}{\scriptsize$\pm$0.07}                                        \\
                Pose Delta                        
                 & 1.10{\scriptsize$\pm$0.15}

                 & 1.13{\scriptsize$\pm$0.15}

                 & 1.19{\scriptsize$\pm$0.16}

                 & \textbf{1.26}{\scriptsize$\pm$0.13}                                        \\
                Pose Margin Accuracy \hspace{0.29em}[\%]
                 & 63.61{\scriptsize$\pm$12.17}

                 & 65.56{\scriptsize$\pm$12.37}

                 & 65.56{\scriptsize$\pm$12.41}

                 & \textbf{71.94}{\scriptsize$\pm$11.38}                                      \\
                \midrule
                Accuracy \hspace{0.29em}[\%]
                 & 73.39{\scriptsize$\pm$5.60}

                & 74.06{\scriptsize$\pm$5.03}
                 & 76.07{\scriptsize$\pm$5.16}

                 & \textbf{77.33}	{\scriptsize$\pm$4.81}                                       \\
                Diag. Acc.\hspace{0.29em}[\%]
                 & 81.94{\scriptsize$\pm$3.32}

                & 83.50{\scriptsize$\pm$4.13}
                 & 84.65{\scriptsize$\pm$3.67}

                 & \textbf{84.93}{\scriptsize$\pm$4.55}                                       \\
                \hspace{0.25em} Sens.\hspace{0.29em}[\%]
                 & 83.14{\scriptsize$\pm$3.71}

                & 84.36{\scriptsize$\pm$4.19}
                 & \textbf{85.50}{\scriptsize$\pm$4.12}

                 & 84.64{\scriptsize$\pm$5.40}                                                \\
                \hspace{0.25em} Spec.\hspace{0.29em}[\%]
                 & 79.56{\scriptsize$\pm$5.83}
                 & 81.78{\scriptsize$\pm$5.51}
                
                 & 82.94{\scriptsize$\pm$8.44}

                 & \textbf{85.50}{\scriptsize$\pm$6.57} \\
            \bottomrule
        \end{tabular}
        }
      \end{minipage}
      \end{table*}
  
      \begin{table*}[htb]
          \centering
          \begin{minipage}[h]{\linewidth}
              \centering
              \caption{Results of the experiments without pretraining, with pretraining on ImageNet, and pretraining on the synthetic dataset and subsequent linear probing on the anatomical preparations dataset. 
              During pretraining with the synthetic data, all augmentations were used.
              Note that the features of the synthetic dataset learned in pretraining have a positive effect on performance during linear probing compared to from-scratch features and the ones learned with ImageNet.}
              \label{table:linear-probing} 
              {
  
        \begin{tabular}{@{}lr|r|cc@{}}
          \toprule
          \multirow{2}{*}{\makecell{Metric}}       
              & \multirow{2}{*}{\makecell{from\\scratch}} 
              & \multirow{2}{*}{\makecell{pretrained\\on ImageNet}}
              & \multicolumn{2}{c}{pretrained on synthetic dataset} \\
          & & & unified camera view & camera view specific \\ 
          \midrule
          MAE                                      
          & 0.58{\scriptsize$\pm$0.00}     
          & 0.55{\scriptsize$\pm$0.02}           
          & \textbf{0.52}{\scriptsize$\pm$0.09}           
          & 0.54{\scriptsize$\pm$0.08}           \\
      Correlation \(r_s\)                      
          & 0.00{\scriptsize$\pm$0.00}     
          & 0.28{\scriptsize$\pm$0.16}           
          & \textbf{0.36}{\scriptsize$\pm$0.29}           
          & 0.35{\scriptsize$\pm$0.31}           \\ 
      \midrule
      Accuracy \hspace{0.29em}[\%]             
          & 42.76{\scriptsize$\pm$0.53}    
          & 47.10{\scriptsize$\pm$2.49}          
          & \textbf{53.61}{\scriptsize$\pm$9.44}          
          & 50.18{\scriptsize$\pm$6.33}          \\
      Diag. Acc.\hspace{0.29em}[\%]            
          & 64.37{\scriptsize$\pm$0.00}    
          & 67.38{\scriptsize$\pm$1.97}          
          & \textbf{69.21}{\scriptsize$\pm$6.90}          
          & 66.39{\scriptsize$\pm$9.17}          \\
      \hspace{0.25em} Sens.\hspace{0.29em}[\%] 
          & 0.00{\scriptsize$\pm$0.00}     
          & 9.15{\scriptsize$\pm$6.07}           
          & \textbf{28.06}{\scriptsize$\pm$20.26}         
          & 27.04{\scriptsize$\pm$24.34}         \\
      \hspace{0.25em} Spec.\hspace{0.29em}[\%] 
          & \textbf{100.0}{\scriptsize$\pm$0.00}    
          & \textbf{100.0}{\scriptsize$\pm$0.00}          
          & 91.0{\scriptsize$\pm$12.95}          
          & 87.49{\scriptsize$\pm$22.09}         \\ 
      \bottomrule
      \end{tabular}
      }
    \end{minipage}
    \end{table*}

\section{Results}
\label{results}
The high accuracy of 87.6\% in Experiment 1 based on the synthetic dataset (see \Cref{table:synthetic-dataset}) shows that synthetic depth images can be used to learn to assess poses and that training with augmented synthetic depth images results in an improvement in nearly all metrics compared to training without augmentation. 
Compared to the diagnostic accuracy of just 59.3\% in the case of a simple baseline that only predicts the mean of all labels, the proposed approach leads to a significant improvement.
Note that the best results are achieved when training with size augmentation.
Since the results of size augmentation are even better than all augmentations combined, and bump augmentation is in fact slightly worse in diagnostic accuracy than those without any augmentation, it can be assumed that bump augmentation has a negative effect on performance in the synthetic dataset.
This may be because the bumps in particular vary greatly in size and position and mask features that are potentially important for pose estimation, such as the ankle.

The results in \Cref{table:anatomical-preparation} for Experiment 2a show that pretraining with the novel synthetic data improves performance on real data.
Compared with training from scratch, ImageNet pretraining does not provide a consistent benefit across the clinically most relevant downstream metrics in this setting.
This suggests that the features learned on ImageNet are not useful for the task of patient pose assessment. 
However, pretraining on the novel synthetic depth images is useful, since almost all metrics are improved over those obtained without pretraining.
Most importantly, the accuracy increases by 11 pp to 90.45\% for the \textit{camera view specific approach} without augmentation.
With the same model, the diagnostic accuracy can also be improved by 3 pp. 

Moreover, the \textit{camera view specific approach} is performing slightly better compared to the \textit{unified camera view approach}  according to the more important accuracy metrics, which is presumably due to the higher similarity of the pretraining and finetuning dataset. 
In contrast to the results from \Cref{table:synthetic-dataset}, the augmentation of the synthetic depth images does not yield any consistent benefits in these experiments.
This is possibly due to an insufficient variance in the shape of the rather few anatomical preparations.
Note however, that bump augmentation in particular has a rather negative impact on performance, which can also be explained by the rather slim and healthy anatomical preparations.

Note that we here obtain an overall accuracy of 90.45\% when predicting diagnostic quality based on depth images. When comparing with the accuracy (93.0\%) obtained for predicting the same diagnostic quality by using real radiographs \citep{mairhoferAIbasedFrameworkDiagnostic2021}, we can conclude that a similar quality assessment is possible with only depth images, although it is significantly more difficult.

To evaluate whether these results can also be transferred to a dataset that contains more subjects and is significantly more diverse and closer to reality than the anatomical preparation dataset, in Experiment 2b we trained on the real-world clinical dataset from scratch and compared it with models that were pretrained on the synthetic dataset and fine-tuned on the real-world clinical dataset.
Since \emph{camera view specific} pretraining performed better when finetuning on the anatomical preparations dataset on average, and size augmentation performed best when training with the synthetic dataset, we also performed \emph{camera view specific} pretraining using size and all augmentations for comparison. 
The results are shown in \Cref{table:real-world}.
Note that the results are considerably worse than the results of the anatomical preparations dataset, both by training from scratch and pretraining with the synthetic dataset.
This discrepancy is due to the lack of ground truth labels. 
Training with poses that are good but not actual 1.0s, or with poses that are bad but not actual 3.0s, can lead to problems. 
This can also affect the metrics after testing, if the pose was actually predicted correctly, but the given label is inaccurate.
Furthermore, movements of the subject within the 5 sampled frames, which are also mapped to the same label, can also lead to inaccuracies.
However, these results still support the claim that pretraining on the synthetic dataset can improve performance compared to training from scratch.
Compared with pretraining on ImageNet, both synthetic-pretraining variants perform better in nearly all reported metrics, with synthetic pretraining using all augmentations yielding the strongest overall results.
The improvement is more apparent in the metrics that measure the ranking of poses rather than the comparison of the prediction with the weak labels.
Thus, we achieve an improvement of approximately 8 pp in pose margin accuracy compared to training from scratch. 
For the accuracy, the difference is only about 4 pp.
Furthermore, for pretraining, it is advantageous with this dataset to train with all augmentations, including bumps, in contrast to previous experiments.  
This is likely due to the high diversity of the subjects in the finetuning test set, who did not suffer from any swelling or edema, but cover both genders and were quite different in proportion, so that in addition to the increased amount of data, the variety of the pretraining data including bumps presumably also had a positive influence on performance.
  
In order to more accurately assess the usefulness of the features learned in pretraining with the synthetic dataset, we finally performed linear probing using the anatomical preparation dataset in Experiment 2c and compared it with a randomly initialized model and a model pretrained on ImageNet.
We also distinguished between the two camera view options in the pretraining \textendash{} both of which were trained with all augmentations.
The results are shown in \Cref{table:linear-probing}.
While the model initialized randomly only predicts the mean value of the labels of the dataset and thus outputs the trivial solution, as can be seen from the sensitivity of 0\% and the specificity of 100\% of the diagnostic accuracy, this experiment also clearly shows that the features learned in pretraining have a positive influence on linear probing.
The \emph{unified camera view} performs best here, which is probably because the pretraining dataset size is twice as large as with the \emph{camera view specific} approach.
However, both approaches outperform a model trained on ImageNet in most metrics.
This final experiment thus further confirms the usefulness of the features obtained from the synthetic dataset.

\subsection{Statistical Analysis}
To further support our main claim that pretraining on synthetic data improves downstream performance on real data beyond the reported mean and standard deviation values, we performed statistical significance tests for the most relevant experiments in this work.
We therefore focused on Experiment 2a (see \Cref{table:anatomical-preparation}) and Experiment 2b (see \Cref{table:real-world}).
By contrast, Experiment 1 primarily serves as an exploratory augmentation ablation study solely on the synthetic dataset and as a basis for selecting later pretraining configuration, whereas Experiment 2c is a supportive analysis of feature usefulness rather than a primary downstream performance comparison.
For all significance tests, the null hypothesis was that pretraining on the synthetic dataset does not improve MAE compared with training from scratch.
Statistical significance was assessed using a one-sided test at a significance level of $\alpha = 0.05$. 

For Experiment 2a, we tested the difference between training from scratch and the \emph{camera view specific} synthetic-pretraining variant with size augmentation, as this was the practically relevant pretraining configuration used in the later experiments and motivated by the previous augmentation study of Experiment 1.
We chose MAE as the only test metric because it directly reflects the regression task, does not depend on any thresholds, and allowed us to evaluate the main claim without introducing additional multiple testing across several correlated metrics.
Since our question is whether pretraining on the synthetic dataset improves performance robustly across repeated runs on this fixed anatomical preparations dataset, we used the 10 repeated runs with different seeds as the independent observations for the statistical test.
For each seed, the MAE values of the two reciprocal preparation folds (see \Cref{experiment-2}) were first averaged, yielding one value per seed and method. These paired seed-level values were then compared using an exact paired permutation test, implemented as a sign-flip test on the paired differences.
Across the paired repeated runs, pretraining on the synthetic dataset reduced the MAE by an average of 0.060 compared with training from scratch.
This effect was significant in the one-sided test ($p = 0.0049$).

For Experiment 2b, we analogously tested the comparison between training from scratch and the synthetic-pretraining variant with all augmentations, since this was the best-performing synthetic-pretraining configuration on the real-world clinical dataset.
Here, we also used MAE as the only test metric for the reasons stated before and to remain consistent with the statistical analysis of Experiment 2a.
As for Experiment 2a, significance was assessed across the 10 repeated runs with different seeds using the same exact paired permutation test.
The only difference is that, for Experiment 2b, the split-level MAE values resulting from the three-fold cross-validation on the real-world clinical dataset (see \Cref{experiment-2}) were first averaged within each seed across the six evaluation folds, rather than across the two reciprocal preparation folds as in Experiment 2a. 
Since these split-level results are components of the same fixed cross-validation evaluation and are not independent observations, they were not tested separately.
Across the paired repeated runs, pretraining on the synthetic dataset reduced the MAE by an average of 0.062 compared with training from scratch.
This effect was significant in the one-sided test ($p = 0.0020$).

Overall, this analysis shows that the main performance gains from pretraining on the synthetic dataset in Experiments 2a and 2b are statistically significant at the predefined 5\% significance level, with both observed p values even below 1\%.

\section{Conclusion, Limitations and Outlook}
In this paper, we presented a framework for generating both synthetic depth images and radiographs from CT scans. 
We have extensively shown that by pretraining on such a synthetic dataset relevant features can be learned, which are useful for the assessment of patients' poses on real data. 
This may help to assess whether the patient's pose would lead to a radiograph with inadequate diagnostic quality before it is taken and thus protect the patient from unnecessary radiation due to a retake.   

The main advantage of this framework is that the data acquisition problem, which is often critical in the medical context, can be solved by using already available CT scans.
Furthermore, the framework can be used to investigate which camera positions and how many cameras are best suited for the pose assessment task \citep{lauferEvaluationofToFpositioning}.
In principle, the framework can also be adapted to different X-ray rooms and ToF camera configurations in order to generate realistic and case-specific synthetic training data.

However, these broader adaptation possibilities remain the subject of future work.
The current validation is restricted to the upper ankle joint and to a single X-ray room with one specific acquisition setup.
In addition, the generated and acquired datasets are relatively small and originate from a single hospital site, which limits the anatomical variability and subject diversity represented in this study.
Furthermore, since the real-world clinical dataset is only weakly labeled and no ground truth radiographs were acquired for it, a prospective clinical study is necessary to evaluate the practical usefulness of the approach, in particular with regard to reducing retakes and unnecessary radiation exposure.

Further future work should therefore evaluate the framework on additional target anatomies, generate larger and more diverse synthetic datasets, and investigate further acquisition environments, including different X-ray rooms, devices, and camera configurations.
Beyond the pose-assessment application demonstrated here, the proposed framework may also be useful for other applications requiring synthetic radiographs and depth images.

\acks{This work was funded by the Bundesministerium für Wirtschaft und Klimaschutz (BMWK) through the KI-SIGS project. We would also like to thank all participants who contributed to the generation of the real-world clinical dataset.}

%
\ethics{The work follows appropriate ethical standards in conducting research and writing the manuscript, following all applicable laws and regulations regarding treatment of animals or human subjects.}

\coi{We declare we don't have conflicts of interest.}

\data{All of the datasets mentioned and used in this work are publicly accessible, or have already been published by us. Our datasets are accessible under \url{https://github.com/INB-KI-SIGS/patient-pose-assessment.git}.} 

\bibliography{bibliography}


\clearpage
\appendix
\makeatletter
\@ifundefined{theHchapter}{}{%
  \renewcommand*{\theHchapter}{app.\Alph{chapter}}%
}
\@ifundefined{theHsection}{}{%
  \renewcommand*{\theHsection}{app.\thesection}%
}
\makeatother

\section{Metrics Definition}
\label{metrics-definition}
The \textbf{Mean Squared Error} (MSE) and \textbf{Mean Absolute Error} (MAE) are calculated as 
\begin{equation}
\label{mse-def}
MSE = \frac{1}{N} \sum_{j=1}^{N}  (y_j - x_j)^2
\end{equation}
and
\begin{equation}
\label{mae-def}
MAE = \frac{1}{N} \sum_{j=1}^{N} \left| y_j - x_j \right|
\end{equation}
where $N$ is the total number of samples, $y_j$ is the prediction, and $x_j$ is the true label.
The \textbf{Spearman correlation coefficient} ($r_s$) is calculated as 
\begin{equation}
\label{rs-def}
    r_s = 1 - \frac{6 \sum_{j=1}^{N} d_j^2}{N(N^2 - 1)}
\end{equation}
where $N$ is the total number of samples, $d_j=R[x_{j}]-R[y_j] $, and $R(x_j)$ is the rank of $x_j$ and $R(y_j)$ is the rank of $y_j$.

\section{Misalignment Example }
\label{misalignment-pc-plane}
\begin{figure}[h]
    \centering
    \includegraphics[width=\linewidth]{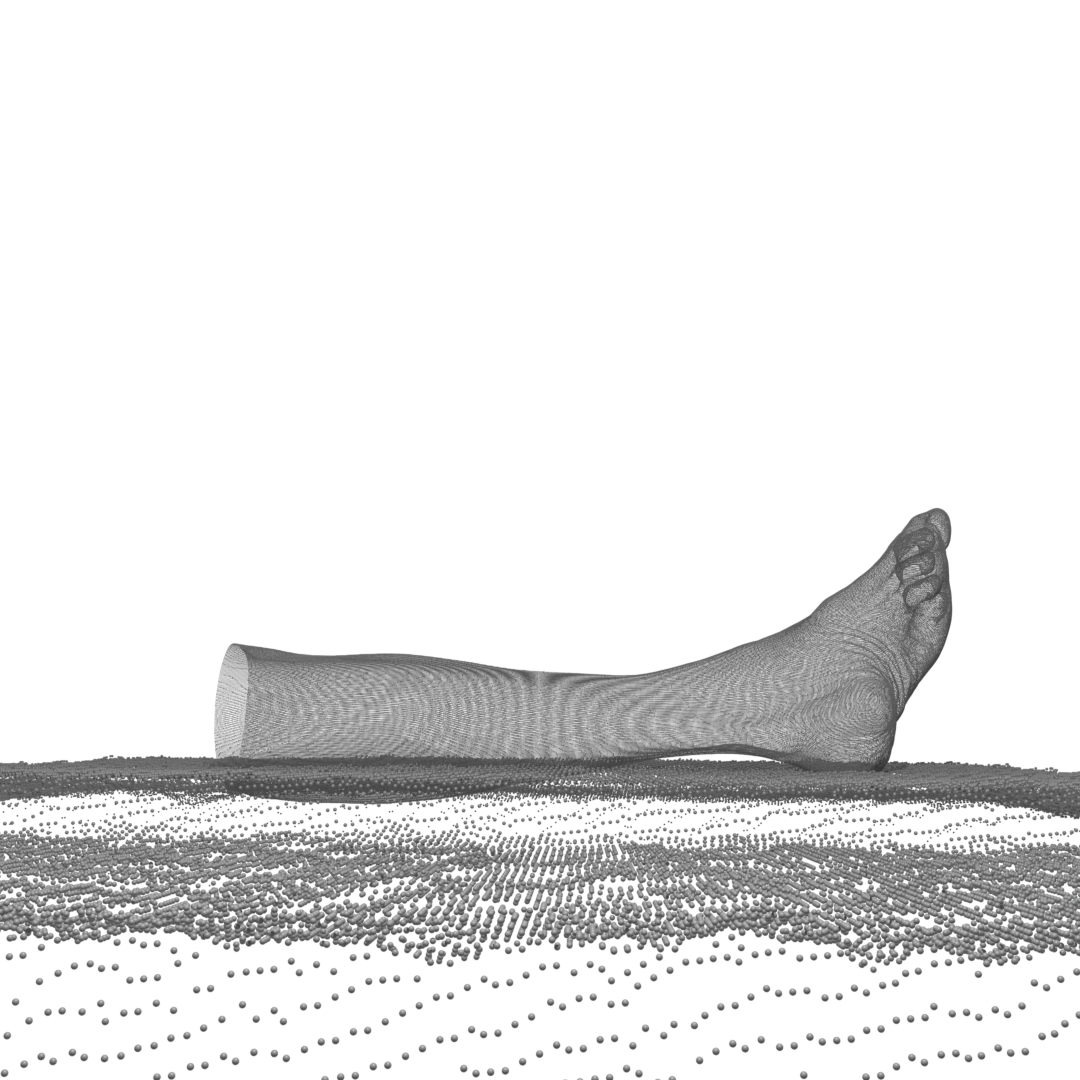}
    \caption{This figure shows a case in which the rotation $T_{rot}$ of the point cloud of the target anatomy and the subsequent application of the matching transformation $T_{mat}$ result in the point cloud not lying realistically on the plane of the table $\Pi_T$. The optimization step needs to be performed.} 
    \label{fig:misalignment-pc-plane}
\end{figure}


\clearpage
\onecolumn
\section{Framework Illustrations}

\label{framework-renderings}
\begin{figure}[htbp]
    \centering
    
    {
        \subfigure[]{%
            \label{fig:rotaion-axis}
            \includegraphics[width=0.4\linewidth]{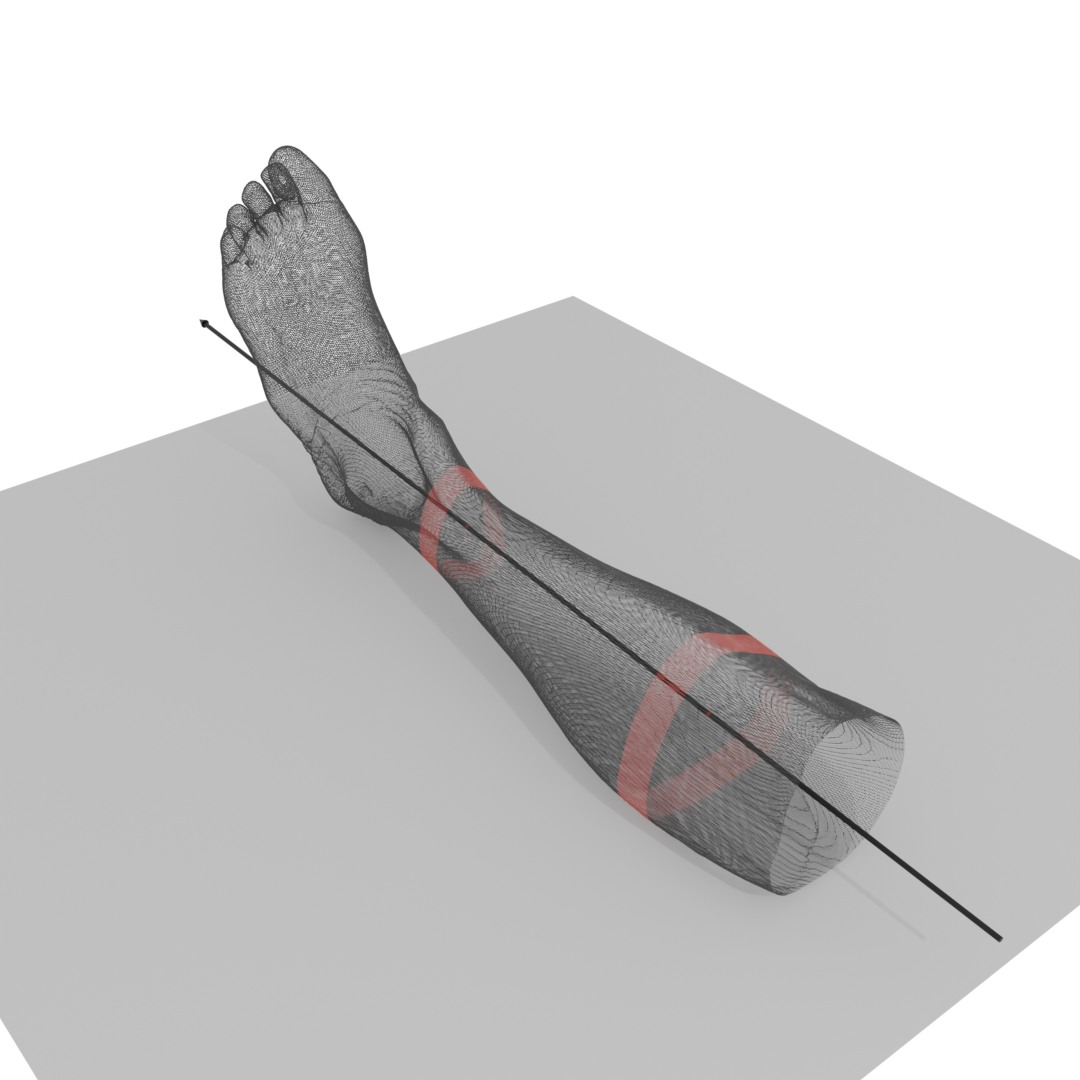}
        }\hfill
        \subfigure[]{%
            \label{fig:pc-augmentation}
            \includegraphics[width=0.4\linewidth]{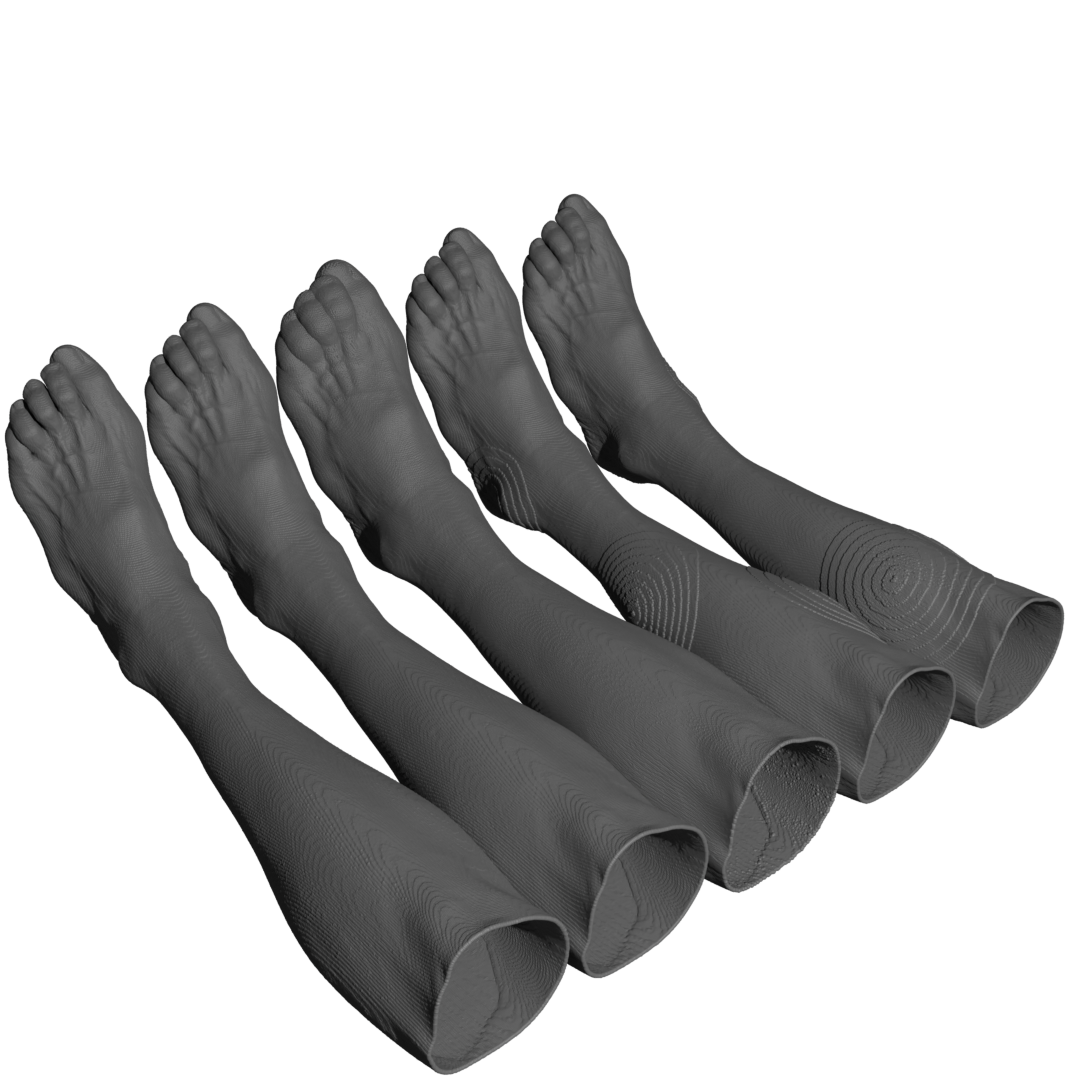}
        }
        \subfigure[]{%
            \label{fig:xray-placement}
            \includegraphics[width=0.4\linewidth]{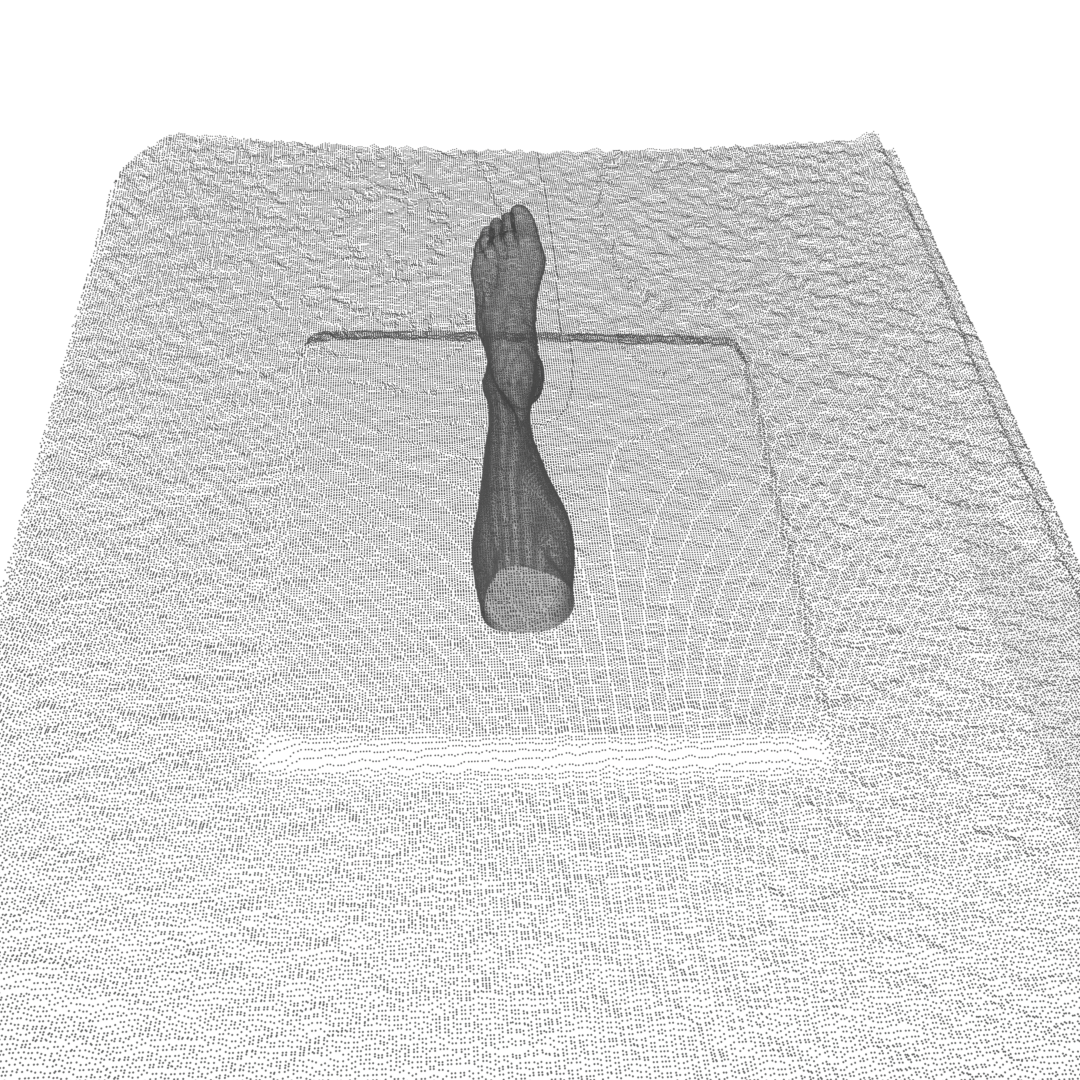}
        }\hfill
        \subfigure[]{
        \label{fig:xray-pc}
        \includegraphics[width=0.4\linewidth]{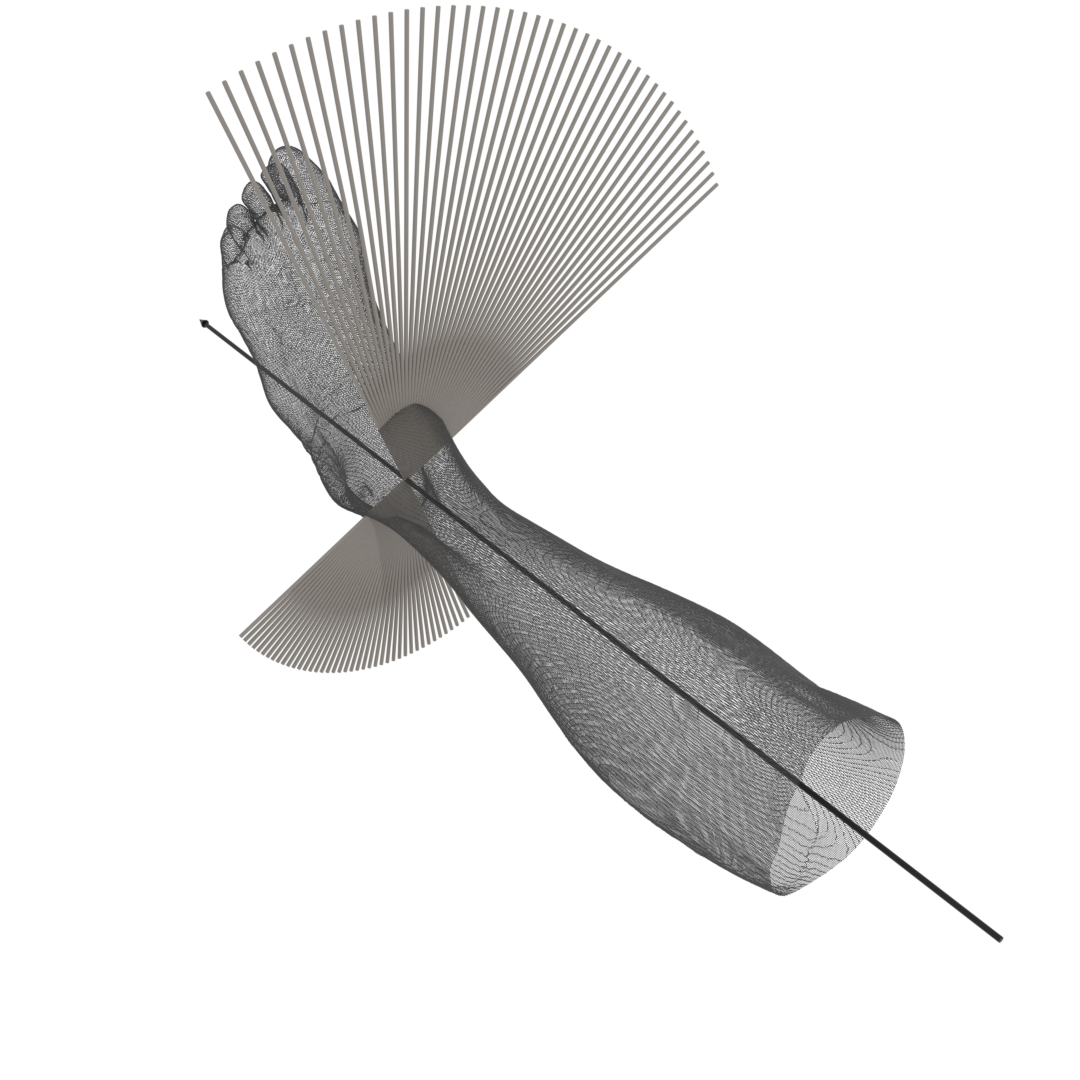}
        }
        \caption{
            This figure illustrates the different steps described in \Cref{framework}.
            \Cref{fig:rotaion-axis} shows the target anatomy cut out of the CT scan as a point cloud and the rotation axis, which is defined by the centers of the points marked in red and passes through the center of the upper ankle joint. The plane $\Pi_P$ is generated below the target anatomy.  
            \Cref{fig:pc-augmentation} shows the augmentation of the point clouds by moving the points along the direction of the normal vectors and adding bumps. 
            For better visibility, the point thickness was increased for this image.
            \Cref{fig:xray-placement} shows the target anatomy combined with the previously acquired X-ray room including the imaging table and detector.
            \Cref{fig:xray-pc} sketches the positions of the X-ray device, which change due to the medial rotation around the longitudinal axis of rotation.
        }
    }
\end{figure}
\twocolumn

\clearpage
\onecolumn
\section{Synthetic Dataset Examples}
\label{example-images}
\begin{figure}[htbp]
    
    \includegraphics[width=\linewidth]{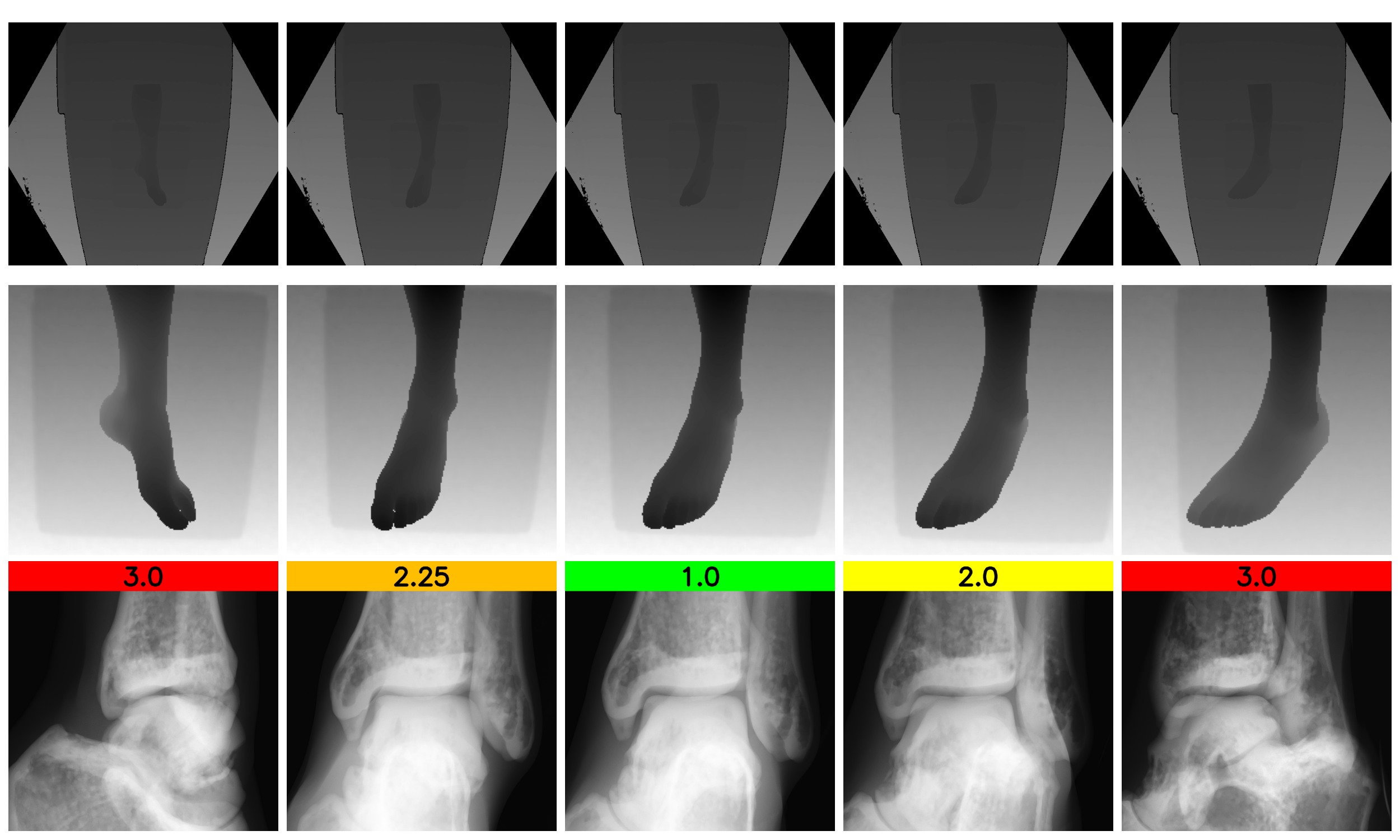}
    \caption{This figure shows synthetic images generated by using the framework. The first row shows the synthetic depth images that were generated with different rotations of the target anatomy.
    The second row shows the manually created ROIs for the synthetic depth images from the first row, which are then used for training. The third row shows the synthetic radiographs corresponding to the synthetic depth images, including their diagnostic quality, which has been assessed by the radiologists. Note that only small rotations are necessary to change a diagnostic quality of 1 to a diagnostic quality of 2, which is reflected in the visibility of the joint space in the different images.}
    \label{fig:synthetic-images}
  \end{figure}

\twocolumn

\clearpage
\onecolumn
\section{Real-World Clinical Dataset Examples}
\label{real-world-examples}

\begin{figure}[htbp]
\centering

\hspace*{0.06\linewidth}%
\makebox[0.30\linewidth][c]{\textbf{(a)}}\hfill
\makebox[0.30\linewidth][c]{\textbf{(b)}}\hfill
\makebox[0.30\linewidth][c]{\textbf{(c)}}\\[1mm]

\newcommand{\rowlbl}[1]{\raisebox{9\height}{\makebox[0.06\linewidth][c]{\textbf{#1}}}}

\rowlbl{1}%
\subfigure{\includegraphics[width=0.30\linewidth]{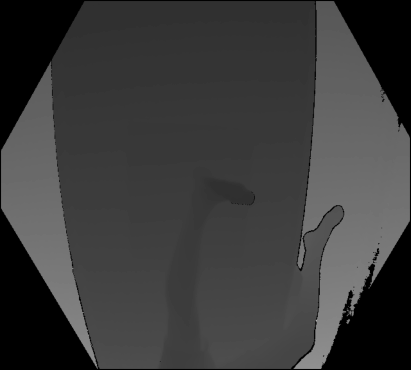}}\hfill
\subfigure{\includegraphics[width=0.30\linewidth]{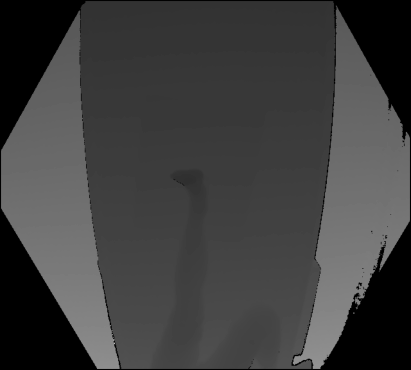}}\hfill
\subfigure{\includegraphics[width=0.30\linewidth]{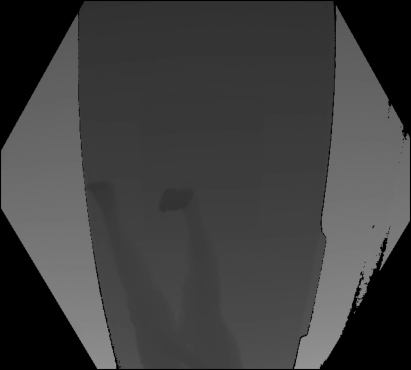}}\\[2mm]

\rowlbl{2}%
\subfigure{\includegraphics[width=0.30\linewidth]{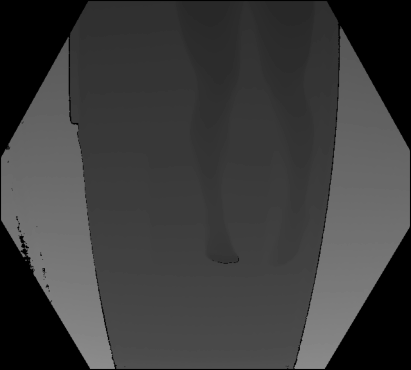}}\hfill
\subfigure{\includegraphics[width=0.30\linewidth]{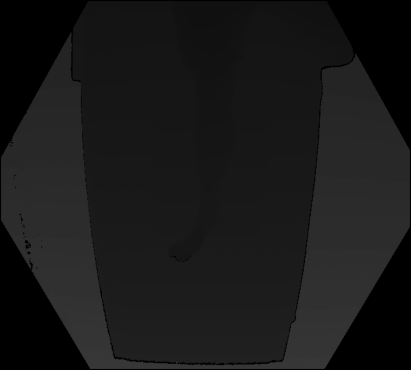}}\hfill
\subfigure{\includegraphics[width=0.30\linewidth]{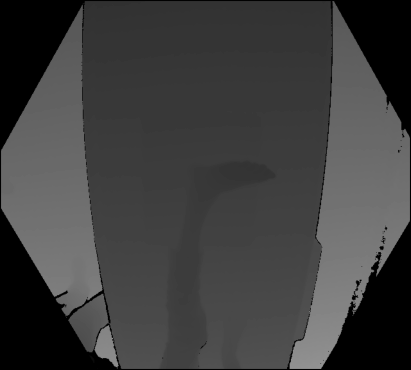}}\\[2mm]

\rowlbl{3}%
\subfigure{\includegraphics[width=0.30\linewidth]{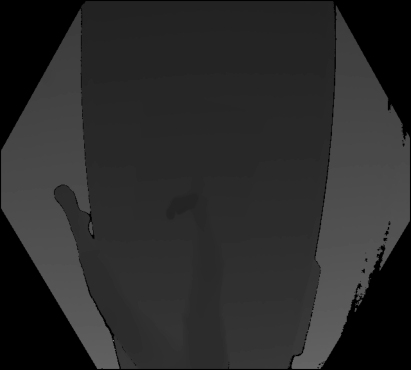}}\hfill
\subfigure{\includegraphics[width=0.30\linewidth]{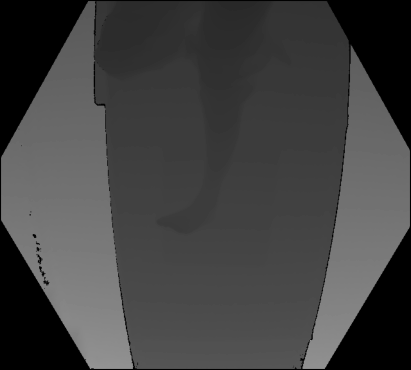}}\hfill
\subfigure{\includegraphics[width=0.30\linewidth]{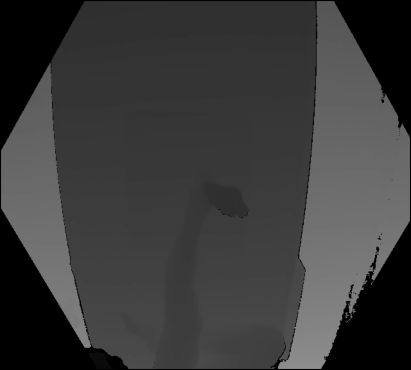}}

\caption{This figure shows examples of the recorded real-world clinical dataset from different subjects from both camera views. Column (a) shows poses that were positioned according to the textbook. Columns (b) and (c) show deliberately bad positioned poses. Note the diversity of the data, for example in terms of the subjects and camera distance and position.}
\label{fig:real-world-images}
\end{figure}
\twocolumn
\clearpage




\end{document}